\documentclass[sigconf]{acmart}

\AtBeginDocument{%
  }

\setcopyright{acmlicensed}
\copyrightyear{2018}
\acmYear{2018}
\acmDOI{XXXXXXX.XXXXXXX}
\acmConference[Conference acronym 'XX]{Make sure to enter the correct
  conference title from your rights confirmation email}{June 03--05,
  2018}{Woodstock, NY}
\acmISBN{978-1-4503-XXXX-X/2018/06}

\usepackage{bm}
\usepackage{algorithm,algorithmic}
\usepackage{multirow}
\usepackage{subcaption}
\usepackage{diagbox}

\begin{document}

\title{DiffGED: Computing Graph Edit Distance via Diffusion-based Graph Matching}

\author{Wei Huang}
\affiliation{%
  \institution{University of New South Wales}
  \country{Australia}
}
\email{w.c.huang@unsw.edu.au}

\author{Hanchen Wang}
\affiliation{%
  \institution{University of Technology Sydney}
  \country{Australia}
}
\email{Hanchen.Wang@uts.edu.au}

\author{Dong Wen}
\affiliation{%
  \institution{University of New South Wales}
  \country{Australia}
}
\email{dong.wen@unsw.edu.au}

\author{Wenjie Zhang}
\affiliation{%
  \institution{University of New South Wales}
  \country{Australia}
}
\email{wenjie.zhang@unsw.edu.au}

\author{Ying Zhang}
\affiliation{%
  \institution{Zhejiang Gongshang University}
  \country{China}
}
\email{ying.zhang@zjgsu.edu.cn}

\author{Xuemin Lin}
\affiliation{%
  \institution{Shanghai Jiaotong University}
  \country{China}
}
\email{xuemin.lin@sjtu.edu.cn}


\begin{abstract}
  The Graph Edit Distance (GED) problem, which aims to compute the minimum number of edit operations required to transform one graph into another, is a fundamental challenge in graph analysis with wide-ranging applications. However, due to its NP-hard nature, traditional A* approaches often suffer from scalability issue, making them computationally intractable for large graphs. Many recent deep learning frameworks address GED by formulating it as a regression task, which, while efficient, fails to recover the edit path—a central interest in GED. Furthermore, recent hybrid approaches that combine deep learning with traditional methods to recover the edit path often yield poor solution quality. These methods also struggle to generate candidate solutions in parallel, resulting in increased running times. 
    
    In this paper, we present a novel approach, DiffGED, that leverages generative diffusion model to solve GED and recover the corresponding edit path. Specifically, we first generate multiple diverse node matching matrices in parallel through a diffusion-based graph matching model. 
    Next, node mappings are extracted from each generated matching matrices in parallel, and each extracted node mapping can be simply transformed into an edit path.
    Benefiting from the generative diversity provided by the diffusion model, DiffGED is less likely to fall into local sub-optimal solutions, thereby achieving superior overall solution quality close to the exact solution. 
    Experimental results on real-world datasets demonstrate that DiffGED can generate multiple diverse edit paths with exceptionally high accuracy comparable to exact solutions while maintaining a running time shorter than most of hybrid approaches.
\end{abstract}

\ccsdesc[500]{Mathematics of computing~Graph algorithms}
\ccsdesc[500]{Mathematics of computing~Combinatorial optimization}
\ccsdesc[500]{Computing methodologies~Machine learning}

\keywords{Graph edit distance, Diffusion model, Graph neural network}


\maketitle

\section{Introduction}
\begin{figure}[t]
     \centering
     \hspace*{-0.5cm}
     \includegraphics[scale=1]{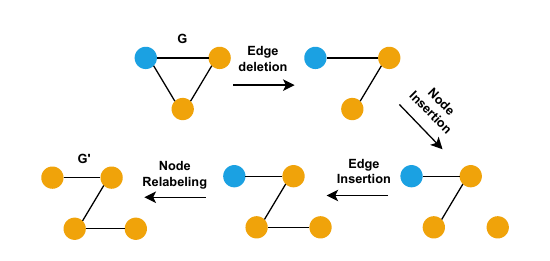}
     \caption{The optimal edit paths for transforming $G$ to $G'$. GED$(G,G')=4$.}
     \label{fig:example}
\end{figure}

Graph edit distance (GED) computation is a fundamental NP-hard problem in graph theory \cite{bunke1997relation}, and GED is also one of the most popular similarity measurements for graphs \cite{gouda2015improved,liang2017similarity}, 
with broad applications in computer vision and pattern recognition, such as scene graph edition \cite{chen2020graph}, image matching \cite{cho2013learning}, and signature verification \cite{maergner2019combining}.
 It aims to determine the minimum number of edit operations required to transform one graph into another as illustrated in Figure \ref{fig:example}. Traditional solvers are mostly designed to find the solution based on A* search \cite{neuhaus2006fast,blumenthal2020exact,chang2020speeding}. However, these solvers often fail to scale to graphs with more than $16$ nodes within reasonable time since the search space grows exponentially with the number of nodes \cite{blumenthal2020exact}. 
In recent years, there has been increasing attention on adopting Graph Neural Networks (GNNs) for GED estimation \cite{bai2019simgnn,bai2021tagsim,zhuo2022efficient,ling2021multilevel,bai2020learning,zhang2021h2mn,qin2021slow}. These approaches typically take a pair of graphs as input and directly estimate GED in one shot through neural networks with extremely short running time. However, the estimated GED might be smaller than the exact GED, which means there is no actual edit path for the estimated GED. 
What is worse, these approaches are not designed to recover the edit path for the estimated GED, where the edit path is often the central interest of many applications \cite{wang2021combinatorial}. 
To overcome these limitations, many hybrid approaches \cite{wang2021combinatorial, yang2021noah} proposed to guide the A* search with an effective and efficient heuristic learned by GNNs.
Unfortunately, these A*-based approaches still suffer from exponential time costs. 

The state-of-the-art approach GEDGNN \cite{piao2023computing} proposed to predict a node matching matrix across two graphs by GNNs, then top-$k$ node mappings with maximum weights are extracted from the predicted node matching matrix to derive the candidate edit paths.
However, the extracted top-$k$ node mappings depend solely on a single node matching matrix and are highly correlated, thus the following limitations could arise:
    (1) The extracted top-$k$ node mappings might fall into the local sub-optimal if the predicted node matching matrix is biased; 
    (2) Highly correlated node mappings limit the diversity of found edit paths, as multiple diverse edit paths could exist with multimodal distribution for an optimal GED; 
    (3) The extraction of top-$k$ node mappings cannot be parallelized to reduce the running time.

In this work, we propose DiffGED, a novel method that utilizes a diffusion-based graph matching model to compute GED with extremely high accuracy and recover its corresponding edit path. 
Specifically, DiffGED first generates $k$ diverse node matching matrices in parallel by our diffusion-based graph matching model DiffMatch. 
Next, $k$ edit paths can be derived by extracting one node mapping from each node matching matrix in parallel using a greedy algorithm, the derived edit path with the minimum number of edit operations will be selected as our solution. 
Therefore, our proposed DiffGED reduces the correlation between each extracted node mapping, which not only enhances overall accuracy and decreases the likelihood of the extracted candidate solutions being locally sub-optimal, but also improves the diversity of the found edit paths.
More importantly, these top-$k$ mapping computations could be parallelized to reduce the running time. 
Our main contributions can be  summarized as follows:
\begin{itemize}
    \item We propose a novel deep learning framework, DiffGED, that computes the edit path for GED by generating multiple node matching matrices and extracting multiple node mappings in parallel.
    \item To the best of our knowledge, this is the first work that proposes a generative diffusion model for graph matching, namely DiffMatch.
    DiffMatch can generate diverse and high-quality node matching matrices and enable the parallelization of top-$k$ node mappings computation. 
    \item Extensive experiments on real-world datasets demonstrate that our proposed DiffGED (1) has exceptionally high accuracy (around $95\%$ on all datasets) which outperforms the baseline methods by a great margin, (2) has great interpretability by generating diverse edit paths, and (3) has a shorter running time compared to other deep learning-based approaches.
    
\end{itemize}

\section{Related Work}
\subsection{Traditional Approaches for GED}
Traditional exact approaches are often based on A* search \cite{blumenthal2020exact,chang2020speeding} guided with designed heuristics to prune unpromising search space. Unfortunately, these exact solvers are usually intractable for large graphs due to the NP-hard nature of GED computation.

To enhance the scalability, traditional approximation approaches focus on constructing a node edition cost matrix, then model GED as a bipartite node matching problem and solve by either Hungarian \cite{riesen2009approximate} or Volgenant-Jonker \cite{bunke2011speeding} algorithm in polynomial time. Another type of approximation method simplifies and accelerates A* search, such as A*-beam search \cite{neuhaus2006fast}, which limits the size of the heap to obtain sub-optimal results in a shorter running time. However, the solution quality of these methods are often poor.
\subsection{Deep Learning Approaches for GED}
To overcome the limitations of traditional approaches, deep learning approaches have been extensively studied in recent years due to the great success of Graph Neural Networks (GNNs) in capturing complex graph structures and solving graph-related tasks. SimGNN \cite{bai2019simgnn} was the first to formulate GED as a regression task, and proposed a cross-graph module to effectively capture relationships between two graphs. It can predict accurate GED within very short running time, inspiring numerous subsequent works \cite{bai2021tagsim,zhuo2022efficient,ling2021multilevel,bai2020learning,zhang2021h2mn,qin2021slow}. However, these approaches cannot recover the edit path, where the edit path is often the central interest of GED.

Another line of research has focused on hybrid approaches that combine deep learning techniques with traditional methods to recover the edit path. Noah \cite{yang2021noah} proposed using a pre-trained Graph Path Network (GPN) as the heuristic for A* beam search. Similarly, GENN-A* \cite{wang2021combinatorial} introduced a Graph Edit Neural Network (GENN) to guide A* search by dynamically predicting the edit costs of unmatched subgraphs. MATA* \cite{liu2023mata} proposed to prune the search space of A* search by extracting top-$k$ candidate matches for each node from two predicted node matching matrices. However, these methods still face scalability challenges similar to those encountered by A* search. To address this, GEDGNN \cite{piao2023computing} adopts a similar approach to VJ and the Hungarian method, where a GNN is used to predict a node matching matrix, reformulating GED as a bipartite node matching problem to improve scalability.

\subsection{Graph Matching}
Graph matching is a problem closely related to GED and deep-learning based graph matching has garnered significant attention across various domains, particularly in image feature matching \cite{jiang2022graph,wang2023deep,chen2019progressive,10.1016/j.patcog.2021.108167}. However, a fundamental distinction between the two problems lies in the nature of their ground truth. In graph matching, the ground truth is typically unique and application-specific, whereas in GED, multiple valid ground truths may exist due to different possible edit paths leading to the same graph transformation.
Additionally, while graph matching focuses on maximizing node correspondence with respect to a predefined ground truth, GED aims to determine the minimal sequence of edit operations required to transform one graph into another. Another key difference lies in the characteristics of the input graphs. In graph matching, the input graphs are often structurally similar, whereas in GED, they can differ significantly. As a result, existing graph matching methods struggle to perform well in GED computation.

\subsection{Graph Similarity}
Graph edit distance is one of the most flexible and expressive graph similarity measures, as it provides a well-defined cost associated with transforming one graph into another through a sequence of edit operations. However, because GED evaluates similarity based on the global structure of graphs, it is often computationally expensive. Beyond GED, an alternative approach is offered by the maximum common subgraph (MCS) \cite{bunke1998graph}, which measures similarity by identifying the largest subgraph common to both graphs. Although MCS is also NP-hard, its computational complexity is typically lower than GED in practice. Specifically, MCS computes graph similarity based on the local substructure and only requires partial node mapping, which allows the search space to be narrowed using anchor nodes.
\begin{algorithm}[t]
    \caption{Edit Path Generation}
    \renewcommand{\algorithmicrequire}{\textbf{Input:}}
    \renewcommand{\algorithmicensure}{\textbf{Output:}}
    \begin{algorithmic}[1]
        \label{algo:edit_path}
        \REQUIRE $G=(V,E,L)$, $G'=(V',E',L')$, node mapping $f$;
        \STATE $EditCost \leftarrow 0$;
        \FOR{each $v \in V$}
            \IF {$L(v) \neq L'(f(v))$}
                \STATE $L(v) \leftarrow L'(v')$;
                \STATE $EditCost \leftarrow EditCost + 1$;
            \ENDIF
        \ENDFOR
        \FOR{each $v' \in V' \setminus \{f(v) \mid v \in V\}$}
            \STATE Create a new $v$;
            \STATE $f(v) \leftarrow v'$ and $L(v) \leftarrow L'(v')$;
            \STATE $V \leftarrow V \cup \{v\}$;
            \STATE $EditCost \leftarrow EditCost + 1$;
        \ENDFOR
        \FOR{each $(v,u) \in E$}
            \IF {$(f(v),f(u)) \in E'$}
                \STATE $E \leftarrow E \setminus \{(v,u)\}$;
                \STATE $EditCost \leftarrow EditCost + 1$;
            \ENDIF
        \ENDFOR
        \FOR{each $(v',u') \in E'$}
            \IF{$(f^{-1}(v),f^{-1}(u))\notin E$}
                \STATE $E \leftarrow E \cup \{(f^{-1}(v),f^{-1}(u))\}$;
                \STATE $EditCost \leftarrow EditCost + 1$;
            \ENDIF
        \ENDFOR
        \RETURN $EditCost$;
    \end{algorithmic}
\end{algorithm}

\subsection{Deep Learning Approaches for Combinatorial Optimization}
In recent years, deep learning has been successfully applied to a variety of combinatorial optimization problems beyond graph edit distance (GED), including the Traveling Salesman Problem (TSP), Maximum Independent Set (MIS), and Maximum Cut (MaxCut). Methods for addressing these problems can be broadly classified into two categories. The first category \cite{khalil2017learning,bai2021glsearch,bello2016neural,kool2018attention} primarily employs reinforcement learning to iteratively construct solutions in an auto-regressive manner. The second category \cite{fu2021generalize,ye2024glop,qiu2022dimes} predicts an initial solution, often represented as a heatmap, which is subsequently refined using traditional optimization techniques. More recently, generative diffusion models \cite{sun2023difusco,graikos2022diffusion} have been applied with notable success in solving the TSP. However, these approaches are predominantly applied to tasks that differ fundamentally from GED in both their settings and objectives.

\subsection{Diffusion Model}
Diffusion models have emerged as a powerful class of generative models, achieving remarkable success in image generation and setting new benchmarks for high-quality image synthesis \cite{ho2020denoising,dhariwal2021diffusion,sohl2015deep,song2019generative}.These models progressively refine random noise into structured outputs through a learned denoising process, demonstrating superior performance over traditional generative approaches such as GANs and VAEs. The success of diffusion models in continuous domains has inspired extensions to discrete data, leading to the development of discrete diffusion models for structured tasks, such as text generation \cite{austin2021structured}. Building on these advancements, discrete diffusion has been extensively applied to graph generation \cite{vignac2022digress,haefeli2022diffusion}, where it has shown great potential in downstream tasks such as molecule generation, further motivating the exploration of diffusion-based approaches for broader graph-based problems beyond generation.
\begin{figure*}[ht]
     \centering
     \hspace*{-0.5cm}
     \includegraphics[scale=0.68]{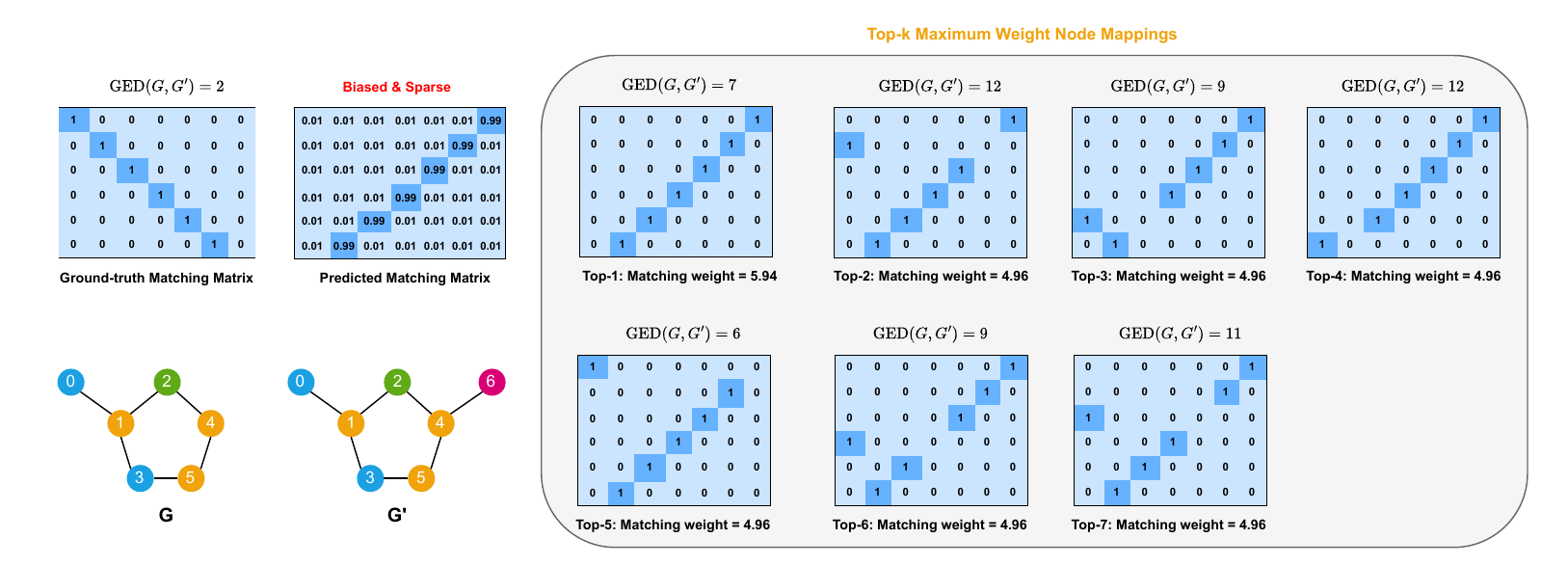}
     \caption{An example of top-$k$ maximum weight node mappings extracted from a biased and sparse predicted node matching matrix.}
     \label{fig:biased_example}
 \end{figure*}

\section{Preliminaries}
\label{sec:preliminary}
In this paper, we focus on the computation of graph edit distance between a pair of undirected labeled graphs $G=(V,E,L)$ and $G'=(V',E',L')$, where $G$ consists of a set of nodes $V$, a set of edges $E$ and a labeling function $L$ that assigns each node a label.

\subsection{Problem Definition}
\label{sec:preliminary_problem_definition}
\textbf{Graph Edit Distance (GED).} \textit{Given a pair of graphs $(G,G')$, find an optimal edit path with minimum number of edit operations that transforms $G$ to $G'$. An edit path is a sequence of edit operations that transforms $G$ to $G'$.
Graph edit distance GED$(G,G')$ is defined as the number of edit operations in the optimal edit path. }

Specifically, there are three types of edit operations: (1) insert or delete a node; (2) insert or delete an edge; (3) replace the label of a node.

\subsection{Edit Path Generation}
\label{sec:preliminary_edit_path_generation}
Suppose $|V| \leq |V'|$, an edit path of transforming $G$ to $G'$ can be obtained from an injective node mapping $f$ from $V$ to $V'$ in linear time complexity $O(|V'| + |E| + |E'|)$ \cite{piao2023computing}, such that $f(v)=v'$, where $v \in V$ and $v' \in V'$. The overall procedure is shown in Algorithm \ref{algo:edit_path}, and can be described as follows:
\begin{enumerate}
    \item For each mapped node pair $f(v)=v'$, if $L(v) \neq L'(v')$, then replace the label of $v$ with $L'(v')$. 
    \item For the remaining unmapped nodes in $V'$, insert $|V'|-|V|$ nodes into $V$. Each inserted node is mapped to and has the same label as an unmapped node in $V'$. 
    \item For any two pairs of mapped nodes $f(v)=v'$ and $f(u)=u'$, if $(u,v) \in E$ and $(u',v') \notin E'$, delete the edge $(u,v)$ from $E$; if $(u,v) \notin E$ and $(u',v') \in E'$, insert the edge $(u,v)$ into $E$. 
\end{enumerate}
Therefore, to find an optimal edit path with minimum number of edit operations, we only have to find an optimal node mapping $f^*$.

\begin{figure*}[ht]
     \centering
     \hspace*{-0.7cm}
     \includegraphics[scale=0.78]{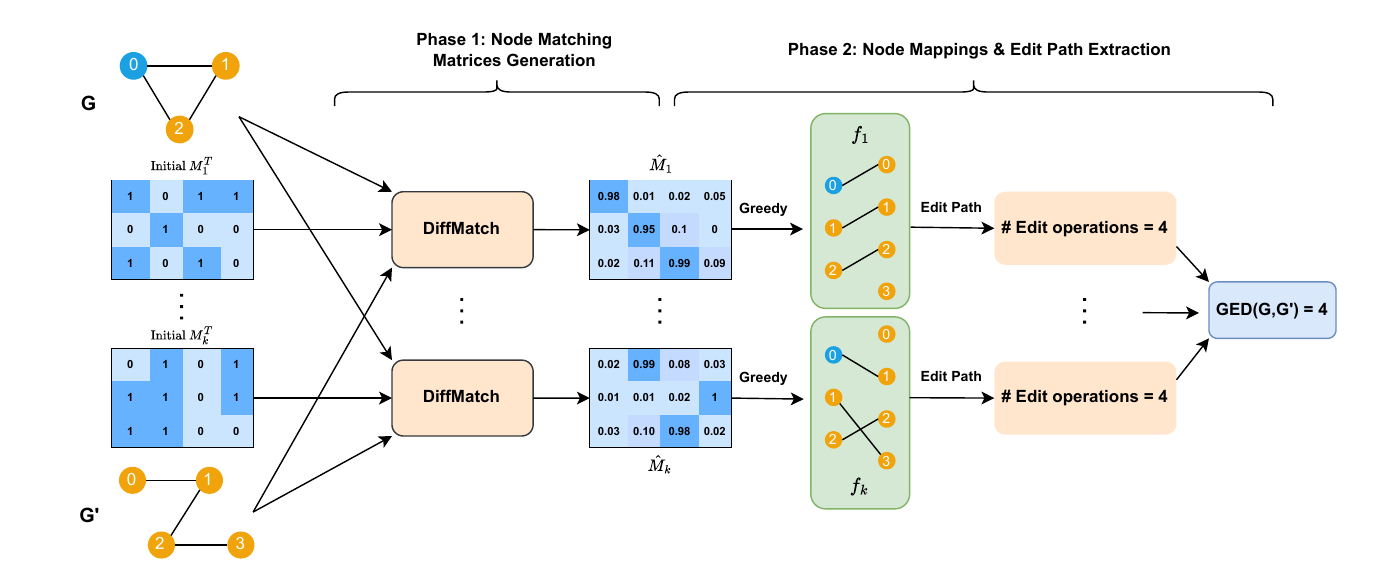}
     \caption{An overview of DiffGED. In the first phase, DiffGED first samples $k$ random initial node matching matrices, then DiffMatch will denoise the sampled node matching matrices. In the second phase, one node mapping will be extracted from each node matching matrix in parallel, and edit paths will be derived from the node mappings.}
     \label{fig:diffged}
 \end{figure*}

 \begin{figure*}[ht]
     \centering
     \hspace*{-0.7cm}
     \includegraphics[scale=0.68]{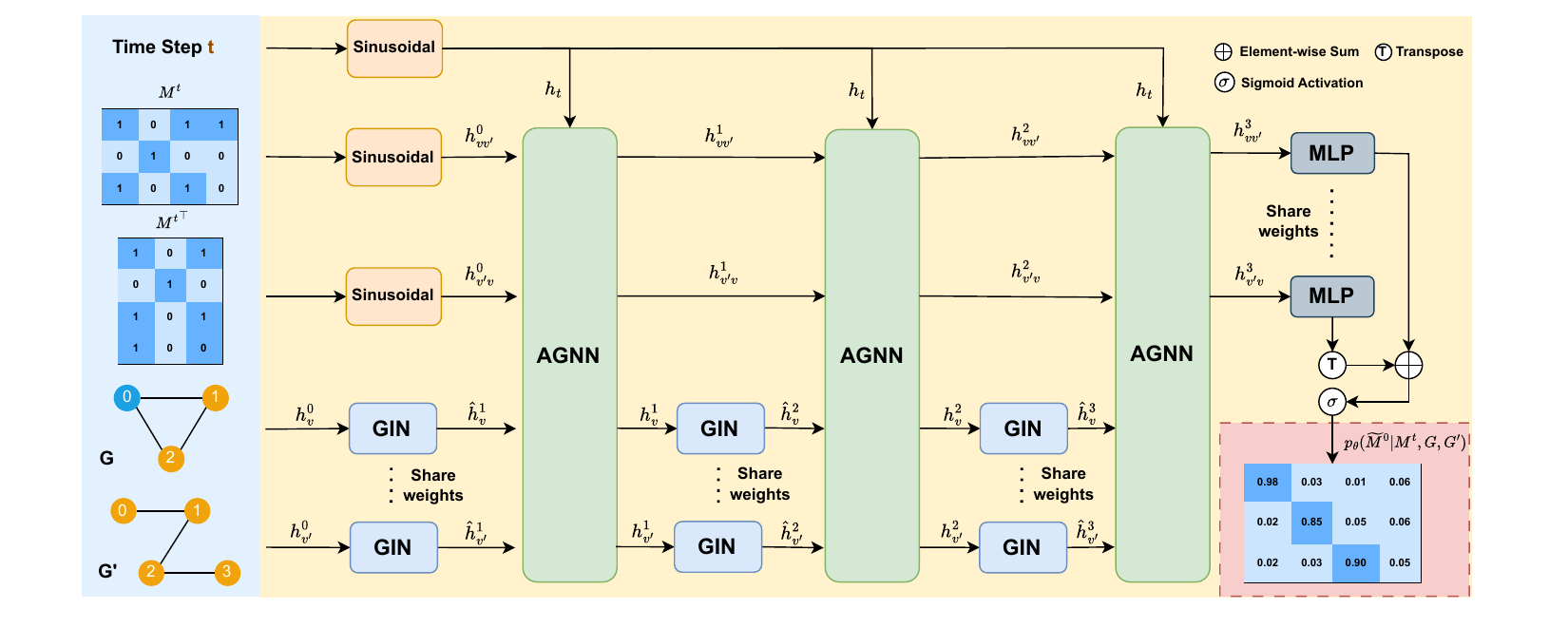}
     \caption{An overview of the denoising network. The blue area denotes the network input, the yellow area denotes the architecture of the denoising network, and the pink area denotes the network output.}
     \label{fig:denoising_network}
 \end{figure*}
 
\section{Proposed Approach}
In this section, we present our DiffGED which leverages the generative diffusion model to predict the optimal edit path.

\subsection{DiffGED: Overview}
\label{sec:overview}
As described in Section \ref{sec:preliminary_edit_path_generation}, the optimal edit path can be obtained from an optimal node mapping $f^*$. To approximately find the optimal node mapping $f^*$, one simple and effective way is to predict top-$k$ node mappings $f_1,...,f_k$, then select the one that results in the edit path with minimum edit operations. 

To generate top-$k$ node mappings, previous works \cite{piao2023computing} focused on a two-phase strategy:
\begin{enumerate}
    \item Predicting a single node matching matrix $\hat{M} \in \mathbb{R}^{|V| \times |V'|}$ by GNNs, where each element $m_{vv'}\in \hat{M}$ represents the weight that node $v \in V$ matches with node $v' \in V'$;
    \item Sequentially extracting top-$k$ node mappings with maximum weights from $\hat{M}$, such that $f_1,...,f_k = Topk(\hat{M})$.
\end{enumerate}
However, extracting from the same node matching matrix will result in $k$ highly correlated node mappings. 
This not only limits the diversity of found edit paths but also makes it prone to falling into local sub-optimal solutions when the predicted node matching matrix is biased, especially when most values in the predicted matrix are similar (sparse) as illustrated in Figure \ref{fig:biased_example}. 
It is clear to see that the top-$k$ node mappings extracted from the predicted matching matrix are highly correlated, and unfortunately, they are all sub-optimal with the derived GED significantly larger than the ground-truth GED. 
Furthermore, due to the sparsity of the predicted matching matrix, the matching weights for the top-$2$ to top-$7$ node mappings are identical, making the extracted top-$k$ mappings uninformative. 
This requires the value of $k$ to be large enough to capture all node mappings with the same matching weight, which increases the computational cost. 
Additionally, the extraction process cannot be parallelized to take advantage of GPU.

Another possible approach is to extract node mappings individually from top-$k$ diverse node matching matrices $\hat{M}_1,...,\hat{M}_k$, such that $f_i = Top1(\hat{M}_i)$, which reduces the correlation between each extracted node mapping, thus decreases the chances of falling into sub-optimal. 
However, the GNNs used in previous works have a limited ability to generate a flexible number of node matching matrices.
Once trained, they can only produce a fixed number of node matching matrices (typically just one), Additionally, this requires a corresponding fixed number of ground-truth node matching matrices, and 
these ground-truth matrices are computationally expensive to obtain.

To address those limitations, our DiffGED leverages the generative diffusion model to generate $k$ different high quality node matching matrices and this $k$ is flexible during inference, independent of the training process, and only requires one ground-truth node matching matrix.
As shown in Figure \ref{fig:diffged}, our DiffGED consists of two phases. 
In the first phase, we sample $k$ random initial discrete node matching matrices $M_i^T \in \{0,1\}^{|V|\times|V'|}$, then iteratively denoise each sampled $M_i^T$ to $\hat{M}_i \in \mathbb{R}^{|V| \times |V'|}$ through the reverse process of our diffusion-based graph matching model DiffMatch. 
In the second phase, we extract one node mapping $f_i$ and generate an edit path from each generated node matching matrix $\hat{M}_i$ efficiently by a simple greedy algorithm in parallel.
The edit path with the minimum edit operations will be our solution.

\subsection{DiffMatch} 

In this sub-section, we introduce our DiffMatch based on a single discrete node matching matrix $M \in \{0,1\}^{|V|\times|V'|}$.

Diffusion models are generative models that consist of a forward process and a reverse process. The forward process $q(M^{1:T}|M^{0})=\prod_{t=1}^{T}q(M^t|M^{t-1})$ progressively corrupts a ground-truth node matching matrix $M^{0}$ to a sequence of increasingly noisy latent variables $M^{1:T}=M^1,M^2,...,M^T$. And the learned reverse process progressively denoises the latent variables towards the desired distribution, starting from a randomly sampled noise $M^T$, the reverse process can be represented as follows:
\begin{equation}
    p_\theta(M^{0:T}|G,G')=p(M^T)\prod_{t=1}^{T}p_\theta(M^{t-1}|M^t,G,G')
\end{equation}

The reason we choose the diffusion model over other generative models is that its reverse process enables the diffusion model to generate node matching matrices through an iterative refinement process, breaking down the complex generation task into simpler steps. Each step makes minor adjustments, progressively improving the quality of the matching matrices. Furthermore, each reverse refinement step will introduce stochasticity, which enhances the model’s ability to produce diverse node matching matrices.

With the great success of discrete diffusion in processing discrete data \cite{haefeli2022diffusion,vignac2022digress,austin2021structured}, we adopt discrete diffusion for DiffMatch.

\subsubsection{Discrete Diffusion Forward Process} 
Let $\widetilde{M}^t \in \{0,1\}^{|V| \times |V'| \times 2}$ be the one-hot encoding of the node matching matrix $M^t$ at time step $t \in [0,T]$.

The forward process adds the noise to the node matching matrix as follow:
\begin{equation}
\begin{aligned}
    q(M^t|M^{t-1})&=\text{Cat}(M^t|p=\widetilde{M}^{t-1}Q_t)\\
    Q_t &= \begin{bmatrix}
                1 - \beta_t & \beta_t \\
                \beta_t & 1 - \beta_t
                \end{bmatrix}
\end{aligned}
\end{equation}
where $Q_t$ is the transition probability matrix, $\beta_t$ is the probability of switching node matching state and Cat denotes the categorical distribution.

To sample the noisy matching matrix $M^t$ efficiently, we can compute the $t$-step marginal from $M^{0}$ as follows:
\begin{equation}
    \begin{aligned}
        q(M^t|M^{0})&=\text{Cat}(M^t|p=\widetilde{M}^{0}\overline{Q}_t) \\
        \overline{Q}_t &= Q_1Q_2...Q_t
    \end{aligned}
\end{equation}

\subsubsection{Discrete Diffusion Reverse Process}
Given a time step $t$, the reverse process denoises the noisy node matching matrix from $M^t$ to $M^{t-1}$, conditioned on the graph pair $G$ and $G'$ as follows:
\begin{equation}
\label{eq:reverse}
    \begin{aligned}
        p_\theta(M^{t-1}|M^t,G,G') &= \sum_{\widetilde{M}} q(M^{t-1}|M^t,\widetilde{M}^{0}) p_\theta(\widetilde{M}^{0}|M^t,G,G') \\
        M^{t-1} &\sim p_\theta(M^{t-1}|M^t,G,G') \\
    \end{aligned}
\end{equation}
where $p_\theta(\widetilde{M}^{0}|M^t,G,G')$ is the node matching probabilities predicted by the denoising network, and the posterior $q(M^{t-1}|M^t,M^{0})$ can be computed as:
\begin{equation}
    \begin{aligned}
        q(M^{t-1}|M^t,M^{0}) =& \frac{q(M^{t}|M^{t-1},M^{0})q(M^{t-1}|M^{0})}{q(M^t|M^{0})}\\
         =& \text{Cat}(M^{t-1};p=\frac{\bar{M}^{t}Q^\top_t \odot \bar{M}^{0} \overline{Q}_{t-1}}{\bar{M}^{0} \overline{Q}_{t} (\bar{M}^{t})^\top})
    \end{aligned}
\end{equation}
where we reshape $\widetilde{M}  \in \{0,1\}^{|V| \times |V'| \times 2}$ to $\bar{M} \in \{0,1\}^{|V||V'| \times 2}$.

\subsubsection{Denoising network} 
The denoising network aims to predict the node matching probabilities $p_\theta(\widetilde{M}^{0}|M^t,G,G')$ given the graph pair and the noisy node matching matrix at time step $t$. Figure \ref{fig:denoising_network} presents an overview of denoising network with 3 layers.

The denoising network takes as input the graph pair$(G,G')$, the time step $t$, and the noisy node matching matrix $M^t$ along with its transpose ${M^t}^\top$. 
Note that, we have GED$(G,G')=$  GED$(G',G)$, therefore we assume symmetry in node matching, meaning if node $v \in V$ matches with node $v' \in V'$, then $v'$ matches with $v$. 
Thus, the forward process is only applied to $M^t \in \mathbb{R}^{|V| \times |V'|}$, while both $M^t$ and ${M^t}^\top$ are taken as inputs to the denoising network for the reverse process. 

Let $\bm{h}^l_v$ and $\bm{h}^l_{v'}$ denote the embedding of node $v \in V$ and $v' \in V'$ at layer $l$, $\bm{h}^l_{vv'}$ and $\bm{h}^l_{v'v}$ denote the embedding of node matching pair $(v,v')$ and $(v',v)$ at layer $l$.
For initialization, the node embeddings $\bm{h}^0_v$ and $\bm{h}^0_{v'}$ are initialized as the one-hot node labels, the node matching pair embeddings $\bm{h}^0_{vv'}$ and $\bm{h}^0_{v'v}$ are initialized as the sinusoidal embeddings \cite{vaswani2017attention} of corresponding values in $M^t$ and ${M^t}^\top$, and the time step embedding $\bm{h}_t$ is initialized as the sinusodial embedding  of $t$.
 
For each layer $l$, the denoising network first updates the node embeddings of each graph to $\hat{\bm{h}}^l_{v}$ and $\hat{\bm{h}}^l_{v'}$, independently using their respective graph structures via GIN \cite{xu2018powerful}. Then, the denoising network further refines the embeddings to ${\bm{h}}^l_{v}$ and ${\bm{h}}^l_{v'}$, while also updating the node matching pair embeddings to ${\bm{h}}^l_{vv'}$ and ${\bm{h}}^l_{v'v}$, by incorporating noisy interactions between node matching pairs and the time step $t$ through Anisotropic Graph Neural Network (AGNN) \cite{joshi2020learning,sun2023difusco,qiu2022dimes}. 

The key advantage of AGNN is its ability to directly compute embeddings for node matching pairs, enabling more expressive representations for cross-graph tasks. In contrast, traditional GNNs such as GIN are specifically designed for computing node embeddings only, making them less suited for capturing relationships between node pairs across graphs. AGNN can be represented as follows:
\begin{equation}
    \begin{aligned}
        \hat{\bm{h}}^l_{vv'} = \bm{W}^l_1 \bm{h}^{l-1}_{vv'}&,\quad \hat{\bm{h}}^l_{v'v} = \bm{W}^l_1 \bm{h}^{l-1}_{v'v}\\
        \tilde{\bm{h}}^l_{vv'} = \bm{W}^l_2 \hat{\bm{h}}^l_{vv'} &+ \bm{W}^l_3 \hat{\bm{h}}^l_{v} + \bm{W}^l_4 \hat{\bm{h}}^l_{v'}\\
        \tilde{\bm{h}}^l_{v'v} = \bm{W}^l_2 \hat{\bm{h}}^l_{v'v} &+ \bm{W}^l_3 \hat{\bm{h}}^l_{v'} + \bm{W}^l_4 \hat{\bm{h}}^l_{v}\\
        {\bm{h}}^l_{vv'} =\hat{\bm{h}}^l_{vv'} + \text{MLP}^l&(\text{ReLU}(\text{GN}_{MM^\top}(\tilde{\bm{h}}^l_{vv'}))+\bm{W}^l_5 \bm{h}_t)\\
        {\bm{h}}^l_{v'v} =\hat{\bm{h}}^l_{v'v} + \text{MLP}^l&(\text{ReLU}(\text{GN}_{MM^\top}(\tilde{\bm{h}}^l_{v'v}))+\bm{W}^l_5 \bm{h}_t)\\
        {\bm{h}}^l_{v} = \hat{\bm{h}}^l_{v} + \text{ReLU}(\text{GN}_{GG'}&(\bm{W}^l_6 \hat{\bm{h}}^l_{v}+ \sum_{v' \in V'}\bm{W}^l_7 \hat{\bm{h}}^l_{v'} \odot \sigma(\tilde{\bm{h}}^l_{vv'})))\\
        {\bm{h}}^l_{v'} = \hat{\bm{h}}^l_{v'} + \text{ReLU}(\text{GN}_{GG'}&(\bm{W}^l_6 \hat{\bm{h}}^l_{v'}+ \sum_{v \in V}\bm{W}^l_7 \hat{\bm{h}}^l_{v} \odot \sigma(\tilde{\bm{h}}^l_{v'v})))
    \end{aligned}
\end{equation}
where $\bm{W}^l_1,\bm{W}^l_2,\bm{W}^l_3,\bm{W}^l_4,\bm{W}^l_5,\bm{W}^l_6,\bm{W}^l_7$ are learnable parameters at layer $l$, $\text{MLP}^l$ denotes multi-layer perceptron at layer $l$, $\text{GN}_{MM^\top}$ is the graph normalization \cite{cai2021graphnorm} over all node matching pairs in both $M^t$ and ${M^t}^\top$, $\text{GN}_{GG'}$ is the graph normalization over all nodes in both $G$ and $G'$, and $\sigma$ is the sigmoid activation. 

Finally, the denoising network computes the matching values of each node pair via multi-layer perceptron (MLP), and sums the matching values for corresponding pairs $(v,v')$ and $(v',v)$, then applies sigmoid activation to obtain the node matching probabilities $p_\theta(\widetilde{M}^{0}|M^t,G,G')$.

\subsubsection{Training of DiffMatch}
 The training procedure of DiffMatch is outlined in Algorithm \ref{algo:training}. For a given graph pair $(G,G')$ sampled from the training data with its ground-truth matching matrix $M^0$, we first sample a time step $t$ from a uniform distribution. Next, we sample a noisy matching matrix $M^t$ from the $t$-step marginal. Finally, the denoising network is trained to minimize the binary cross-entropy loss between the predicted matching matrix $p_\theta(\widetilde{M}^{0}|M^t,G,G')$ and the ground-truth node matching matrix $\widetilde{M}^{0}$.
\begin{algorithm}[t]
    \caption{DiffMatch Training Procedure}
    \renewcommand{\algorithmicrequire}{\textbf{Input:}}
    \begin{algorithmic}[1]
        \label{algo:training}
        \REQUIRE Graph pair $(G,G')$, Ground-truth node matching matrix $M^0$;
        \STATE Sample $t \sim Uniform({1,...,T})$;
        \STATE Sample $M^t \sim q(M^t|M^{0})$;
        \STATE $p_\theta(\widetilde{M}^{0}|M^t,G,G') \leftarrow DenoisingNetwork(G,G',M^t,{M^t}^\top,t)$;
        \STATE Take gradient step on $BCELoss(p_\theta(\widetilde{M}^{0}|M^t,G,G'),M^0)$;
    \end{algorithmic}
\end{algorithm}

\subsubsection{Accelerating DiffMatch Inference}
During training, the forward process typically employs a large number of steps $T$ (e.g., $T=1000$), and performing $T$ reverse steps during inference can be computationally expensive. To accelerate DiffMatch’s inference, we apply DDIM \cite{song2020denoising} to the reverse process. The key idea of DDIM is that, instead of performing $T$ reverse steps over the entire sequence $[T,...,1]$, we perform only $S$ reverse steps on a sub-sequence $[\tau_S,...,\tau_1]$ of $[T,...,1]$, where $S < T$ and $\tau_S = T$. We substitute $t$ and $t-1$ in Equation \ref{eq:reverse} with $\tau_i$ and $\tau_{i-1}$, and we modify the posterior as follows:
\begin{equation}
    \begin{aligned}
        q(M^{\tau_{i-1}}|M^{\tau_i},M^{0}) =& \frac{q(M^{\tau_i}|M^{\tau_{i-1}},M^{0})q(M^{\tau_{i-1}}|M^{0})}{q(M^{\tau_i}|M^{0})}\\
         =& \text{Cat}(M^{\tau_{i-1}};p=\frac{\bar{M}^{\tau_i}\overline{Q}^\top_{{\tau_{i-1}},{\tau_i}} \odot \bar{M}^{0} \overline{Q}_{\tau_{i-1}}}{\bar{M}^{0} \overline{Q}_{\tau_i} (\bar{M}^{\tau_i})^\top})\\
         \overline{Q}_{{\tau_{i-1}},{\tau_i}} =& Q_{\tau_{i-1}+1}Q_{\tau_{i-1}+2}...Q_{\tau_{i}}
    \end{aligned}
\end{equation}

Algorithm \ref{algo:inference} outlined the reverse process of DiffMatch during inference, given a graph pair with node matching matrix randomly sampled from Bernoulli distribution.
Note that, for the last reverse step, we use $\hat{M} = p_\theta(M^{0}|M^{\tau_1},G,G')$ as the input of the node mappings extraction in phase $2$. 

\begin{algorithm}[t]
    \caption{Sampling from DiffMatch}
    \renewcommand{\algorithmicrequire}{\textbf{Input:}}
    \begin{algorithmic}[1]
        \label{algo:inference}
        \REQUIRE Graph pair $(G,G')$, Random node matching matrix $M^T$;
        \FOR{$\tau_i=\tau_S$ to $\tau_1$}
            \STATE $p_\theta(\widetilde{M}^{0}|M^{\tau_i},G,G') \leftarrow DenoisingNetwork(G,G',M^{\tau_i},{M^{\tau_i}}^\top,\tau_i)$;
            \IF{$\tau_i \neq \tau_1$}
                \STATE $M^{\tau_{i-1}} \sim p_\theta(M^{\tau_{i-1}}|M^{\tau_i},G,G')$;
            \ELSE
                \STATE $\hat{M} \leftarrow p_\theta(M^{0}|M^{\tau_1},G,G')$;
            \ENDIF
        \ENDFOR
        \RETURN $\hat{M}$;
    \end{algorithmic}
\end{algorithm}

\subsubsection{Time complexity Analysis} For a $N$-layer denoising network with a hidden dimension of $d$, the time complexity of GIN within a single layer is $O(|V'|d^2+max(|E|,|E'|)d)$, and the time complexity of AGNN within a single layer is $O(|V||V'|d^2)$. Thus, the overall time complexity of the denoising network is $O(N(max(|E|,|E'|)d + |V||V'|d^2))$, and the overall time complexity of the reverse process with $S$ steps is $O(S(N(max(|E|,|E'|)d + |V||V'|d^2)))$.

\subsection{Node Mapping Extraction}
After sampling $k$ noisy node matching matrices $M^T_1,...,M^T_k$ and denoising to $\hat{M}_1,...,\hat{M}_k$, we adopt the greedy algorithm based on matching weights to extract one node mapping from each node matching matrix as shown in Algorithm \ref{algo:greedy} (assuming $|V| \leq |V'|$). 
\begin{algorithm}[t]
    \caption{Greedy Node Mapping Extraction}
    \renewcommand{\algorithmicrequire}{\textbf{Input:}}
    \renewcommand{\algorithmicensure}{\textbf{Output:}}
    \begin{algorithmic}[1]
        \label{algo:greedy}
        \REQUIRE $i$-th node matching matrix $\hat{M}_{i} \in \mathbb{R}^{|V| \times |V'|}$;
        \ENSURE $i$-th node mapping $f_i$;
        \STATE Initialize $f_i \leftarrow \emptyset$ ;
        \FOR{$n \gets 1$ to $|V|$}
            \STATE select $(v,v')$ with the maximum value in $\hat{M}_{i}$;
            \STATE $f_i \leftarrow f_i \cup \{(v,v')\}$;
            \STATE set all elements in $v$-th row of $\hat{M}_{i}$ to $-\infty$;
            \STATE set all elements in $v'$-th column of $\hat{M}_{i}$ to $-\infty$;
        \ENDFOR
        \RETURN $f_i$;
    \end{algorithmic}
\end{algorithm}
Specifically, the greedy node mapping extraction starts by selecting the node pair with the highest matching probability. Once a node pair is selected, all matching probabilities involving either of the selected nodes are set to $-\infty$ to prevent them from being selected again. This process is repeated iteratively until every node in $V$ is assigned to a corresponding node in $V'$.

Note that, the above greedy algorithm does not guarantee the extraction of optimal node mappings from the node matching matrices, but it has a time complexity of $O(|V|^2|V'|)$ slightly faster than the exact Hungarian algorithm \cite{kuhn1955hungarian} with time complexity of $O(|V'|^3)$. 
It can also be easily parallelized by GPU to extract $k$ node mappings from $k$ node matching matrices simultaneously to reduce the running time, especially for large $k$. It will be demonstrated in Section \ref{sec:ablation} that DiffGED with the above greedy algorithm is sufficient to achieve excellent performance.

\section{Experiments}

\begin{table}[t]
\caption{Dataset description}
    \label{table:dataset}
    \centering
    \begin{tabular}{|c|c|c|c|c|}
    \hline
    Dataset&\# Graphs & Avg $|V|$ &Avg $|E|$&Max $|V|$\\
    \hline
    AIDS700&$700$&$8.9$&$8.8$&$10$\\
    \hline
    Linux&$1000$&$7.6$&$6.9$&$10$ \\
    \hline
    IMDB&$1500$&$13$&$65.9$&$89$ \\
    \hline
    \end{tabular}
\end{table}

\subsection{Dataset}
\label{sec:dataset}
We conduct experiments over three popular real-world GED datasets: AIDS700 \cite{bai2019simgnn}, Linux \cite{wang2012efficient,bai2019simgnn} and IMDB \cite{bai2019simgnn,yanardag2015deep}. Each graph in AIDS700 is labeled, while each graph in Linux and IMDB is unlabeled. The statistics of datasets are summarized in Table \ref{table:dataset}. We obtain the ground-truth edit path (node mappings) from \cite{piao2023computing}. However, the ground-truth GED and edit paths are often computationally expensive to obtain for graph pairs with at least one graph has more than $10$ nodes. To handle this, we follow the same strategy as described in \cite{piao2023computing} to generate synthetic graphs for IMDB dataset. Specifically, for each graph $G$ with more than $10$ nodes, synthetic graphs are generated by randomly applying $\Delta$ edit operations to $G$, these random edit operations are used as an approximation of the ground-truth edit path and $\Delta$ is used as an approximate of ground-truth GED. For graphs with more than $20$ nodes, $\Delta$ is randomly distributed in $[1,10]$, for graphs with more than $10$ nodes and less than $20$ nodes, $\Delta$ is randomly distributed in $[1,5]$.

For each dataset, we split $60\%$, $20\%$, and $20\%$ of all the graphs as training set, validation set, and testing set, respectively. To form training pairs, each training graph with no more than $10$ nodes is paired with all other training graphs with no more than $10$ nodes, each training graph with more than $10$ nodes is paired with $100$ synthetic graphs. 
In the validation and testing sets, each graph with no more than $10$ nodes is paired with $100$ random training graphs with no more than $10$ nodes, and each graph with more than $10$ nodes is paired with $100$ synthetic graphs.

\begin{table*}[ht]
\centering
\caption{Overall performance on testing graph pairs. Methods with a running time exceeding $24$ hours are marked with -.}
\label{tab:result}
\begin{tabular}{|c|c|c|c|c|c|c|c|c|}
\hline
Datasets & Models & MAE & Accuracy & $\rho$ & $\tau$ &p@$10$&p@$20$&Time(s) \\
\hline
\hline
\multirow{7}{*}{AIDS700}&Hungarian&$8.247$&$1.1\%$&$0.547$&$0.431$&$52.8\%$&$59.9\%$&$0.00011$\\
&VJ&$14.085$&$0.6\%$&$0.372$&$0.284$&$41.9\%$&$52\%$&$0.00017$\\
\cline{2-9}
&Noah&$3.057$&$6.6\%$&$0.751$&$0.629$&$74.1\%$&$76.9\%$&$0.6158$\\
&GENN-A*&$0.632$&$61.5\%$&$0.903$&$0.815$&$85.6\%$&$88\%$&$2.98919$\\
&GEDGNN&$1.098$&$52.5\%$&$0.845$&$0.752$&$89.1\%$&$88.3\%$&$0.39448$\\
&MATA*&$0.838$&$58.7\%$&$0.8$&$0.718$&$73.6\%$&$77.6\%$&$\textbf{0.00487}$\\
&DiffGED (ours)&$\textbf{0.022}$&$\textbf{98\%}$&$\textbf{0.996}$&$\textbf{0.992}$&$\textbf{99.8\%}$&$\textbf{99.7\%}$&$0.0763$\\
\hline
\hline
\multirow{7}{*}{Linux}&Hungarian&$5.35$&$7.4\%$&$0.696$&$0.605$&$74.8\%$&$79.6\%$&$0.00009$\\
&VJ&$11.123$&$0.4\%$&$0.594$&$0.5$&$72.8\%$&$76\%$&$0.00013$\\
\cline{2-9}
&Noah&$1.596$&$9\%$&$0.9$&$0.834$&$92.6\%$&$96\%$&$0.24457$\\
&GENN-A*&$0.213$&$89.4\%$&$0.954$&$0.905$&$99.1\%$&$98.1\%$&$0.68176$\\
&GEDGNN&$0.094$&$96.6\%$&$0.979$&$0.969$&$98.9\%$&$99.3\%$&$0.12863$\\
&MATA*&$0.18$&$92.3\%$&$0.937$&$0.893$&$88.5\%$&$91.8\%$&$\textbf{0.00464}$\\
&DiffGED (ours)&$\textbf{0.0}$&$\textbf{100\%}$&$\textbf{1.0}$&$\textbf{1.0}$&$\textbf{100\%}$&$\textbf{100\%}$&$0.06982$\\
\hline
\hline
\multirow{7}{*}{IMDB}&Hungarian&$21.673$&$45.1\%$&$0.778$&$0.716$&$83.8\%$&$81.9\%$&$0.0001$\\
&VJ&$44.078$&$26.5\%$&$0.4$&$0.359$&$60.1\%$&$62\%$&$0.00038$\\
\cline{2-9}
&Noah&-&-&-&-&-&-&-\\
&GENN-A*&-&-&-&-&-&-&-\\
&GEDGNN&$2.469$&$85.5\%$&$0.898$&$0.879$&$92.4\%$&$92.1\%$&$0.42428$\\
&MATA*&-&-&-&-&-&-&-\\
&DiffGED (ours)&$\textbf{0.937}$&$\textbf{94.6\%}$&$\textbf{0.982}$&$\textbf{0.973}$&$\textbf{97.5\%}$&$\textbf{98.3\%}$&$\textbf{0.15105}$\\
\hline

\end{tabular}
\end{table*}

\begin{table*}[ht]
\centering
\caption{Overall performance on unseen testing graph pairs. Methods with a running time exceeding $24$ hours are marked with -.}
\label{tab:result_unseen}
\begin{tabular}{|c|c|c|c|c|c|c|c|c|}
\hline
Datasets & Models & MAE & Accuracy & $\rho$ & $\tau$ &p@$10$&p@$20$&Time(s) \\
\hline
\hline
\multirow{7}{*}{AIDS700}&Hungarian&$8.237$&$1.5\%$&$0.527$&$0.416$&$54.3\%$&$60.3\%$&$0.0001$\\
&VJ&$14.171$&$0.9\%$&$0.391$&$0.302$&$44.9\%$&$52.9\%$&$0.00016$\\
\cline{2-9}
&Noah&$3.174$&$6.8\%$&$0.735$&$0.617$&$77.8\%$&$76.4\%$&$0.5765$\\
&GENN-A*&$0.508$&$67.1\%$&$0.917$&$0.836$&$87.1\%$&$90.6\%$&$3.44326$\\
&GEDGNN&$1.155$&$50.5\%$&$0.838$&$0.746$&$89.1\%$&$87.6\%$&$0.39344$\\
&MATA*&$0.885$&$56.6\%$&$0.77$&$0.689$&$73.2\%$&$76.6\%$&$\textbf{0.00486}$\\
&DiffGED (ours)&$\textbf{0.024}$&$\textbf{96.4\%}$&$\textbf{0.993}$&$\textbf{0.986}$&$\textbf{99.7\%}$&$\textbf{99.7\%}$&$0.07546$\\
\hline
\hline
\multirow{7}{*}{Linux}&Hungarian&$5.423$&$7.5\%$&$0.725$&$0.623$&$75\%$&$77\%$&$0.00008$\\
&VJ&$11.174$&$0.4\%$&$0.613$&$0.512$&$70.6\%$&$74.5\%$&$0.00013$\\
\cline{2-9}
&Noah&$1.879$&$8\%$&$0.872$&$0.796$&$84.3\%$&$92.2\%$&$0.25712$\\
&GENN-A*&$0.142$&$92.9\%$&$0.976$&$0.94$&$99.6\%$&$99.6\%$&$1.17702$\\
&GEDGNN&$0.105$&$96.2\%$&$0.979$&$0.968$&$98.6\%$&$98.5\%$&$0.12169$\\
&MATA*&$0.201$&$91.5\%$&$0.948$&$0.903$&$86.2\%$&$90.2\%$&$\textbf{0.00464}$\\
&DiffGED (ours)&$\textbf{0.0}$&$\textbf{100\%}$&$\textbf{1.0}$&$\textbf{1.0}$&$\textbf{100\%}$&$\textbf{100\%}$&$0.06901$\\
\hline
\hline
\multirow{7}{*}{IMDB}&Hungarian&$21.156$&$45.9\%$&$0.776$&$0.717$&$84.2\%$&$82.1\%$&$0.00012$\\
&VJ&$44.072$&$26.6\%$&$0.4$&$0.359$&$60.1\%$&$63.1\%$&$0.00037$\\
\cline{2-9}
&Noah&-&-&-&-&-&-&-\\
&GENN-A*&-&-&-&-&-&-&-\\
&GEDGNN&$2.484$&$85.5\%$&$0.895$&$0.876$&$92.3\%$&$91.7\%$&$0.42662$\\
&MATA*&-&-&-&-&-&-&-\\
&DiffGED (ours)&$\textbf{0.932}$&$\textbf{94.6\%}$&$\textbf{0.982}$&$\textbf{0.974}$&$\textbf{97.5\%}$&$\textbf{98.4\%}$&$\textbf{0.15107}$\\
\hline

\end{tabular}
\end{table*}

\subsection{Baseline methods}
For traditional approximation methods, we compare our DiffGED with \textbf{Hungarian} \cite{riesen2009approximate} and \textbf{VJ} \cite{bunke2011speeding}. For deep learning methods, we compare with the following hybrid methods that can generate an edit path: (1) \textbf{Noah} \cite{yang2021noah} uses Graph Path Network (GPN) to supervise A*-beam search; (2) \textbf{GENN-A*} \cite{wang2021combinatorial} uses Graph Edit Neural Network (GENN) to guide A* search; (3) \textbf{MATA*} \cite{liu2023mata} generates two node matching matrices, then extracts top-$k$ candidate matching nodes in $G'$ for each node in $G$ to construct the search space, then applies A*LSa \cite{chang2020speeding}; (4) \textbf{GEDGNN} \cite{piao2023computing} generates a single node matching matrix, then extracts top-$k$ node mappings to generate edit paths.

\subsection{Implementation details}
During training of our DiffMatch, we set the number of time steps $T$ to $1,000$ with linear noise schedule, where $\beta_0 = 10^{-4}$ and $\beta_T=0.02$. For the reverse denoising process during testing, we set the number of time steps $S$ to $10$ with linear denoising schedule, and we generate $k=100$ node matching matrices in parallel for each testing graph pair.

For our denoising network, we set the number of layers to $6$, the output dimension of each layer is $128$, $64$, $32$, $32$, $32$, $32$, respectively. We train it for $200$ epochs with batch size of $128$, we adopt Adam optimizer \cite{kingma2014adam} with learning rate of $0.001$ and weight decay of $5\times10^{-4}$.

All experiments are conducted using Nvidia Geforce RTX3090 24GB and Intel i9-12900K with 128GB RAM.

\subsection{Evaluation Metrics}
We evaluate our DiffGED against other baseline methods based on the following metrics: (1) \textbf{\textit{Mean Absolute Error (MAE)}} measures the average absolute difference between the predicted GED and the ground-truth GED; (2) \textbf{\textit{Accuracy}} measures the ratio of the testing graph pairs with predicted GED equals to the ground-truth GED.

For each testing graph $G$, we pair $G$ with another $100$ graphs $G'_1,...,G'_{100}$ to form graph pairs as described in Section \ref{sec:dataset}, we rank the similarity of each $G'_1,...,G'_{100}$ to $G$ based on the ground-truth GED and the predicted GED of $(G,G'_i)$, respectively. We evaluate the ranking results using the following metrics: (1) \textbf{\textit{Spearman’s
Rank Correlation Coefficient ($\rho$)}}, and (2) \textbf{\textit{Kendall’s Rank Correlation
Coefficient ($\tau$)}}, both measure the matching ratio between the ground-truth ranking results and the predicted ranking results; (3) \textbf{\textit{Precision at top-$10$ and top-$20$}} (p@$10$, p@$20$) measure the ratio of predicted top-$10$ and top-$20$ similar graphs within the ground-truth top-$10$ and top-$20$ similar graphs, respectively.

Moreover, we compare the efficiency of each method based on the average running time over all testing pairs.

\subsection{Results}
\label{sec:result}
Table \ref{tab:result} presents the overall performance of all methods on the test pairs. Across all datasets, DiffGED demonstrates exceptionally high solution quality in terms of MAE, accuracy, and all ranking metrics. For the AIDS700 dataset, the accuracy of DiffGED is nearly double that of other hybrid approaches. DiffGED consistently shows shorter running times than most hybrid approaches across all datasets, although it is slower than MATA* on smaller datasets. Note that, all A*-based hybrid approaches fail to complete evaluations on (IMDB) within a reasonable time due to the scalability issues inherent in A* search.

Specifically, both MATA* and DiffGED need to predict node matching matrices and then extract top-$k$ candidate results. However, they differ in key aspects:
(1) MATA* predicts only two node matching matrices in a single step, whereas DiffGED predicts $k$ node matching matrices in parallel over $10$ denoising steps. This results in faster node matching matrix generation for MATA*; (2) MATA* extracts the top-$k$ candidate matching nodes in $G'$ for each node in $G$, limiting the valid range of $k$ to $|V'|$ and typically selecting a small $k$ to reduce the A* search space. In contrast, DiffGED extracts the top-$k$ global maximum weight node mappings, allowing $k$ to be arbitrarily large. As a result, MATA* achieves shorter running times on smaller datasets. However, on larger datasets, MATA* suffers from the exponential growth of the A* search space, whereas DiffGED remains unaffected by this limitation.

Moreover, while GEDGNN can scale to large graphs and follows a procedure similar to our DiffGED, it is slower and performs worse across all datasets for several reasons. GEDGNN sequentially extracts top-$k$ candidate node mappings from a single node matching matrix, resulting in highly correlated mappings. In contrast, DiffGED extracts top-$k$ candidate node mappings from $k$ node matching matrices in parallel, generating diverse mappings. This diversity reduces the likelihood of falling into local sub-optimal solutions, even if some predicted node matching matrices are biased. Additionally, the parallelization of node mapping extraction significantly reduces runtime. 

\subsection{Generalization Ability}
To evaluate the generalization ability to unseen graphs of our DiffGED, instead of pairing each testing graph with $100$ graphs from the training set, we pair each testing graph with $100$ unseen graphs from the testing set. Table \ref{tab:result_unseen} presents the overall performance of all methods on these unseen testing graph pairs. Compared to the results in Table \ref{tab:result}, it demonstrates that DiffGED can still achieve superior performance without losing accuracy, even with more challenging unseen testing graph pairs.

Moreover, in real-world scenarios, obtaining ground-truth node mappings for large graph pairs is often impractical. To evaluate the generalization ability of DiffGED under such conditions, we modify the training setup. Instead of training each method on a combination of real small graph pairs and synthetic large graph pairs from IMDB, we train each method exclusively on real small graph pairs from IMDB. However, the testing set still consists of a combination of real small graph pairs and synthetic large graph pairs.
Table \ref{table:generalization} presents the overall performance of DiffGED and GEDGNN when trained on real small graph pairs. As observed, the accuracy of both DiffGED and GEDGNN degrades, primarily because the testing graph pairs differ from the training graph pairs not only in graph size but also in distribution, due to the presence of synthetic graph pairs in the testing set, as these synthetic graphs differ from real graph pairs. Despite this challenge, DiffGED still outperforms GEDGNN, achieving higher accuracy.

\begin{figure}[htbp]
    \centering
    
    \begin{subfigure}{0.49\linewidth}
        \centering
        \includegraphics[width=\linewidth]{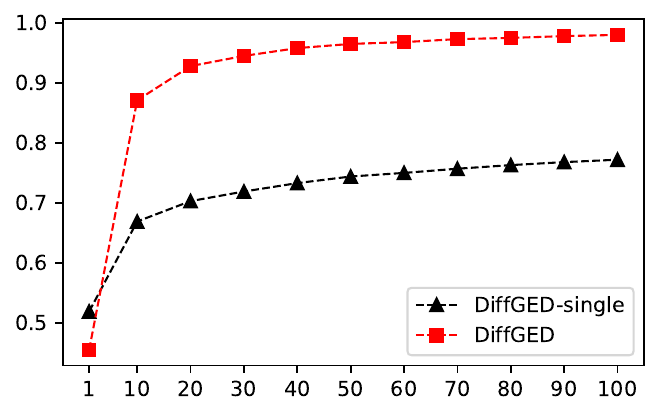}
        \caption{AIDS - Accuracy vs. k}
    \end{subfigure}
    \hfill
    \begin{subfigure}{0.49\linewidth}
        \centering
        \includegraphics[width=\linewidth]{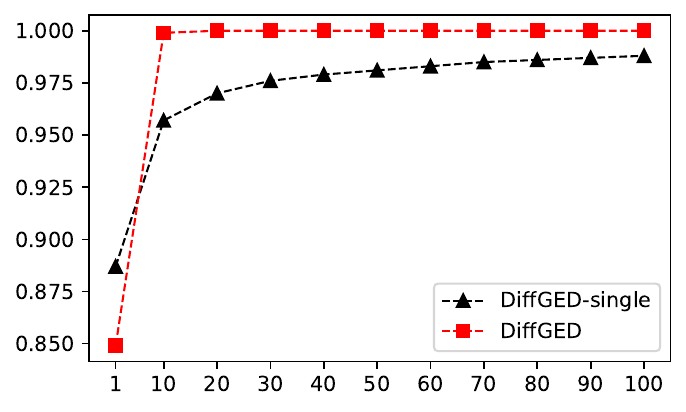}
        \caption{Linux - Accuracy vs. k} 
    \end{subfigure}

    \begin{subfigure}{0.49\linewidth}
        \centering
        \includegraphics[width=\linewidth]{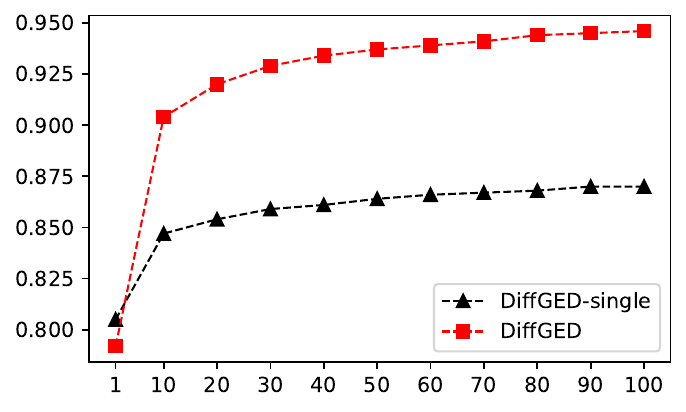}
        \caption{IMDB - Accuracy vs. k} 
    \end{subfigure}
    \hfill
    \begin{subfigure}{0.49\linewidth}
        \centering
        \includegraphics[width=\linewidth]{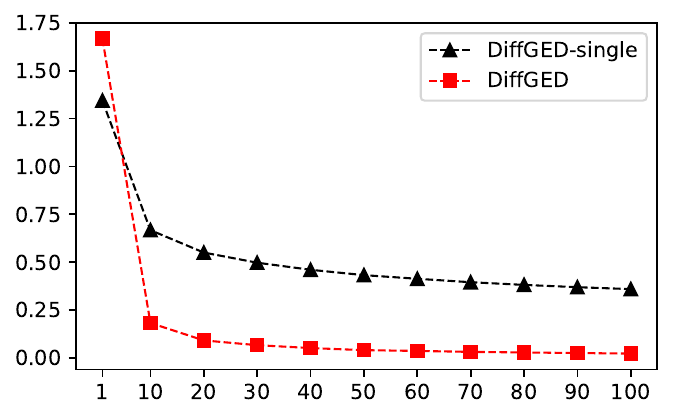}
        \caption{AIDS - MAE vs. k} 
    \end{subfigure}

    \begin{subfigure}{0.49\linewidth}
        \centering
        \includegraphics[width=\linewidth]{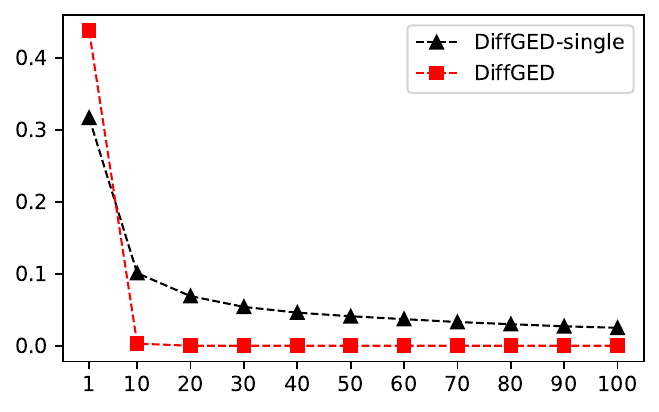}
        \caption{Linux - MAE vs. k} 
    \end{subfigure}
    \hfill
    \begin{subfigure}{0.49\linewidth}
        \centering
        \includegraphics[width=\linewidth]{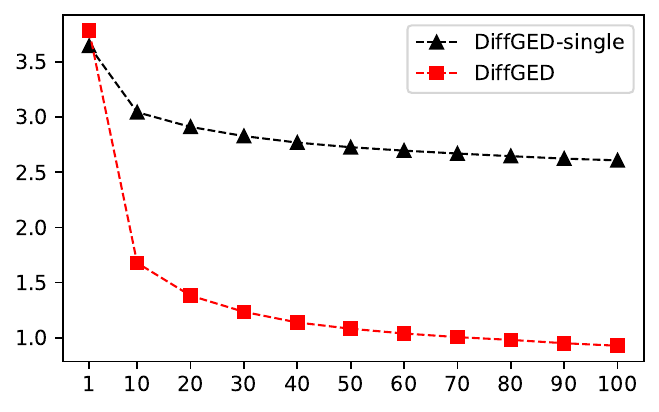}
        \caption{IMDB - MAE vs. k} 
    \end{subfigure}

    \begin{subfigure}{0.49\linewidth}
        \centering
        \includegraphics[width=\linewidth]{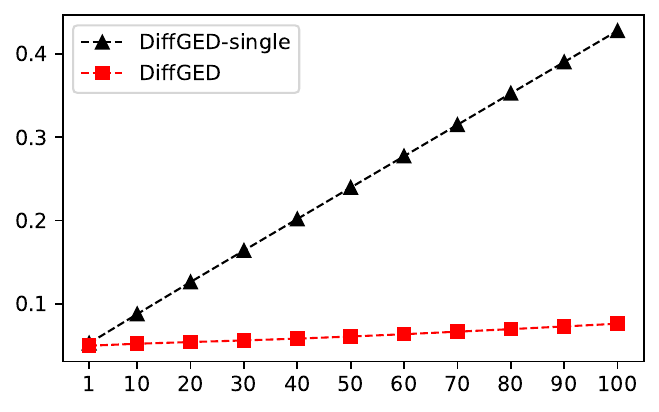}
        \caption{AIDS - Time (s) vs. k} 
    \end{subfigure}
    \hfill
    \begin{subfigure}{0.49\linewidth}
        \centering
        \includegraphics[width=\linewidth]{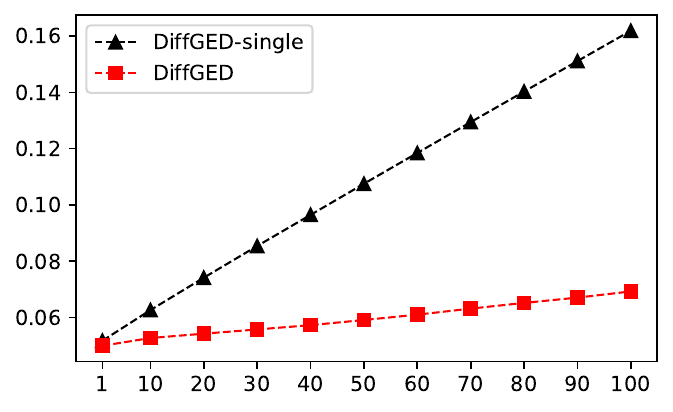}
        \caption{Linux - Time (s) vs. k} 
    \end{subfigure}

    \begin{subfigure}{0.49\linewidth}
        \centering
        \includegraphics[width=\linewidth]{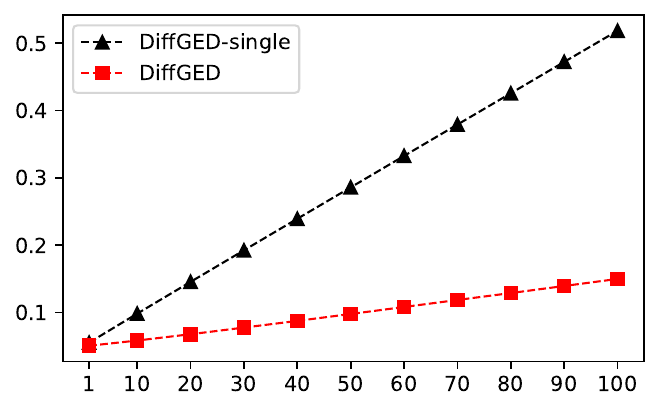}
        \caption{IMDB - Time (s) vs. k} 
    \end{subfigure}

    \caption{Effectiveness and Efficiency of Top-$k$ Approaches}
    \label{fig:topk}
\end{figure}

\begin{table*}[ht]
\centering
\caption{Overall Performance on IMDB testing graph pairs. IMDB-small refers to training set that only contains real small graph pairs. IMDB-mix refers to training set that contains a combination of real small graph pairs and synthetic large graph pairs.}
\label{table:generalization}
\begin{tabular}{|c|c|c|c|c|c|c|c|c|}
\hline
Training set & Models & MAE & Accuracy & $\rho$ & $\tau$ &p@$10$&p@$20$&Time(s) \\
\hline
\hline
\multirow{2}{*}{IMDB-small}&GEDGNN&$7.943$&$77.1\%$&$0.844$&$0.815$&$88.2\%$&$87.6\%$&$0.48253$\\
&DiffGED&$5.789$&$83\%$&$0.892$&$0.874$&$90.1\%$&$90.8\%$&$0.14923$\\
\hline
\hline
\multirow{2}{*}{IMDB-mix}&GEDGNN&$2.469$&$85.5\%$&$0.898$&$0.879$&$92.4\%$&$92.1\%$&$0.42428$\\
&DiffGED&$0.937$&$94.6\%$&$0.982$&$0.973$&$97.5\%$&$98.3\%$&$0.15105$\\
\hline

\end{tabular}
\end{table*}

\subsection{Ablation Study}
\label{sec:ablation}
\noindent\textbf{DiffGED top-$k$ vs GEDGNN top-$k$.}\quad To better evaluate the effectiveness, efficiency and edit path diversity of the top-$k$ node mappings generation in our DiffGED model compared to the approach proposed by GEDGNN, we create a variant model, DiffGED-single. This variant generates only a single node matching matrix using DiffMatch and then applies the top-$k$ extraction method proposed in GEDGNN. 

As illustrated in Figure \ref{fig:topk}(a)-(f), our top-$k$ approach (DiffGED) performs slightly worse than GEDGNN (DiffGED-single) when $k=1$. This difference arises because GEDGNN uses the exact Hungarian algorithm for top-1 node mapping, while DiffGED employs an approximate greedy strategy. However, as $k$ increases, this initial disadvantage diminishes, with DiffGED rapidly converging to near-optimal accuracy and MAE, even with its approximate greedy method. In contrast, DiffGED-single, despite using an exact extraction algorithm, converges to sub-optimal accuracy. Notably, for simpler datasets like Linux, DiffGED achieves optimal solution quality with a small value of $k=10$. The key reason behind this is that DiffGED generates a more diverse set of node mappings, which helps avoid sub-optimal solutions, whereas GEDGNN's mappings tend to be highly correlated, leading to sub-optimal results. Moreover, even with GEDGNN's top-$k$ approach, it is interesting to note that DiffGED-single with $k=100$ still achieves higher accuracy across all datasets compared to the result of GEDGNN in Table \ref{tab:result}, which highlights the effectiveness of our DiffMatch module.

Furthermore, as shown in Figure \ref{fig:topk}(g)-(i), the running time of DiffGED-single increases significantly faster than that of DiffGED as $k$ grows. This disparity arises from DiffGED-single’s sequential top-$k$ node mapping strategy, whereas DiffGED benefits from parallelized node matching matrix generation and parallel node mapping extraction. Since both processes in DiffGED are parallelized, the impact of increasing $k$ on its running time remains minimal, underscoring its superior efficiency for larger $k$ values.

\begin{figure}
    \centering
    \begin{minipage}[b]{0.49\linewidth}
        \centering
        \begin{subfigure}{\linewidth}
            \includegraphics[width=\linewidth]{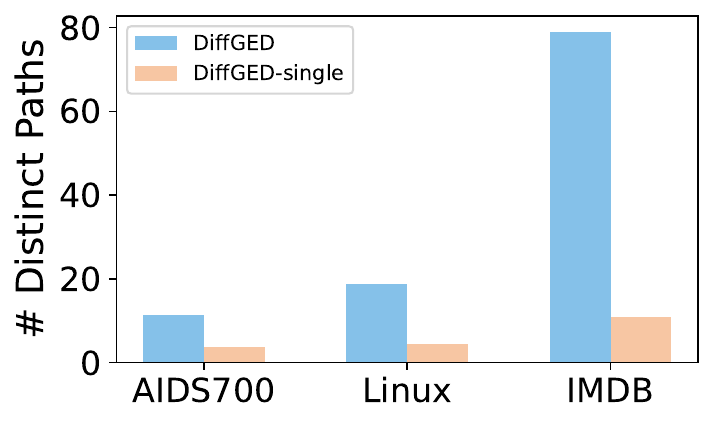}
            \caption{GT-GED}
        \end{subfigure}
    \end{minipage}
    \hfill
    \begin{minipage}[b]{0.49\linewidth}
        \centering
        \begin{subfigure}{\linewidth}
            \includegraphics[width=\linewidth]{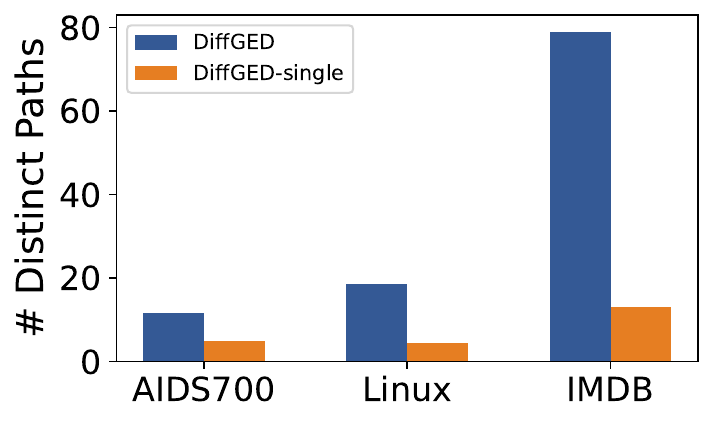}
            \caption{Pred-GED}
        \end{subfigure}
    \end{minipage}
    \caption{Evaluation of Found Edit Path Diversity. Pred-GED refers to average number of distinct edit paths with predicted minimum GED. GT-GED refers to average number of distinct edit paths with ground-truth GED.}
    \label{fig:diversity}
\end{figure}

\begin{table*}[ht]
\centering
\caption{Ablation study on testing graph pairs. }
\label{tab:ablation}
\begin{tabular}{|c|c|c|c|c|c|c|c|c|}
\hline
Datasets & Models & MAE & Accuracy & $\rho$ & $\tau$ &p@$10$&p@$20$&Time(s) \\
\hline
\hline
\multirow{4}{*}{AIDS700}&DiffGED&${0.022}$&${98\%}$&${0.996}$&${0.992}$&${99.8\%}$&${99.7\%}$&$0.0763$\\
&DiffGED(w/o diffusion)&${1.618}$&${46.7\%}$&${0.732}$&${0.629}$&${82.4\%}$&${81.1\%}$&$0.01179$\\
&GEDGNN&$1.098$&$52.5\%$&$0.845$&$0.752$&$89.1\%$&$88.3\%$&$0.39448$\\
&GEDGNN(AGNN)&$0.736$&$66.7\%$&$0.884$&$0.812$&$94\%$&$931\%$&$0.39112$\\
\hline
\hline
\multirow{4}{*}{Linux}
&DiffGED&${0.0}$&${100\%}$&${1.0}$&${1.0}$&${100\%}$&${100\%}$&$0.06982$\\
&DiffGED(w/o diffusion)&${0.743}$&${74.7\%}$&${0.887}$&${0.839}$&${96.4\%}$&${95.8\%}$&$0.01117$\\
&GEDGNN&$0.094$&$96.6\%$&$0.979$&$0.969$&$98.9\%$&$99.3\%$&$0.12863$\\
&GEDGNN(AGNN)&$0.061$&$97.4\%$&$0.992$&$0.987$&$99.6\%$&$99.5\%$&$0.13164$\\
\hline
\hline
\multirow{4}{*}{IMDB}
&DiffGED&${0.937}$&${94.6\%}$&${0.982}$&${0.973}$&${97.5\%}$&${98.3\%}$&${0.15105}$\\
&DiffGED(w/o diffusion)&${0.832}$&${93.3\%}$&${0.942}$&${0.93}$&${98.6\%}$&${96.8\%}$&$0.01944$\\
&GEDGNN&$2.469$&$85.5\%$&$0.898$&$0.879$&$92.4\%$&$92.1\%$&$0.42428$\\
&GEDGNN(AGNN)&$1.766$&$89.1\%$&$0.903$&$0.89$&$93.9\%$&$92.8\%$&$0.41387$\\
\hline

\end{tabular}
\end{table*}
Lastly, we evaluate edit paths diversity by computing the average number of distinct edit paths found per graph pair, where the number of edit operations is equal to the predicted minimum GED and the ground-truth GED, respectively, using $k=100$. As demonstrated in Figure \ref{fig:diversity}, our method is capable of generating multiple distinct edit paths for both the predicted minimum GED and the ground-truth GED, while the top-$k$ approach used in GEDGNN is limited to generating only a few. This is due to the fact that diverse optimal edit paths often exist within a multimodal distribution. Our approach can generate diverse top-$k$ mappings, allowing us to effectively capture this multimodal distribution. In contrast, the approach used by GEDGNN generates highly correlated node mappings towards one mode, which limits its ability to capture the range of possible edit paths.

\vspace{1mm}\noindent \textbf{Do we really need diffusion?} The core idea of the proposed framework is to generate diverse, high-quality node matching matrices through an iterative reverse process of the diffusion model. To assess the effectiveness of the diffusion model in DiffMatch, we introduce a one-shot generative variant model, DiffGED(w/o diffusion), which takes a graph pair and a randomly initialized node matching matrix as input and directly predicts the clean node matching matrix, followed by greedy node mapping extraction. In this setup, we remove the time step component from the denoising network. During training, DiffGED(w/o diffusion) is also provided with a random node matching matrix instead of a noisy node matching matrix sampled from the forward diffusion process. 

Table \ref{tab:ablation} presents the overall performance of DiffGED(w/o diffusion). Notably, DiffGED (w/o diffusion) performs poorly, and its performance is even worse than GEDGNN on the AIDS and Linux datasets.

From a solution quality perspective, DiffGED(w/o diffusion) attempts to generate a high-quality node matching matrix in a single step from random noise, making the learning task extremely challenging. In contrast, the diffusion model decomposes this complex generation task into simpler, iterative refinements. The reverse diffusion process gradually denoises the random node matching matrix step by step, ensuring that each step only requires minor corrections. This progressive refinement leads to higher-quality node matching matrices.

From a solution diversity perspective, DiffGED introduces stochasticity at each reverse step during inference, whereas the stochasticity in DiffGED(w/o diffusion) comes solely from the random noise input. As a result, DiffGED is more likely to generate diverse node matching matrices. Furthermore, in diffusion models, the training input consists of a ground-truth node matching matrix corrupted by the forward diffusion process, rather than pure noise, and noisy matching matrix is only mapped to the ground-truth matching matrix. However, in DiffGED(w/o diffusion), the training input is pure noise, requiring a single random noise to map to multiple 
ground-truth matching matrices. This one-to-many mapping increases the likelihood of mode collapse, reducing the model’s ability to generate diverse solutions. Therefore, diffusion model is necessary for our DiffGED to generate high quality and diverse node matching matrices. But it is interesting to note that the running time of DiffGED (w/o diffusion) is much shorter than DiffGED since it generates node matching matrices in one-shot without iteration.

\vspace{1mm}\noindent \textbf{Anisotropic Graph Neural Network} Instead of computing only node embeddings and then using their inner product to predict node matching probabilities, our denoising network leverages the Anisotropic Graph Neural Network (AGNN) to directly compute node pair embeddings, enabling a more expressive prediction of node matching probabilities. 

To evaluate the effectiveness of AGNN, we create a variant of GEDGNN, GEDGNN(AGNN), that replaces its Cross Matrix Module with AGNN (without time steps). Moreover, we initialize a fixed node matching matrix filled with ones as input of GEDGNN(AGNN). We choose to create a variant of GEDGNN rather than creating a variant of DiffMatch by replacing AGNN with the Cross Matrix Module. This is because DiffMatch requires a noisy node matching matrix as input, but the Cross Matrix Module of GEDGNN (MLP($[h_v^\top W_1h_{v'},...,h_v^\top W_ch_{v'}]$)) cannot incorporate such noisy information when computing node matching probabilities. This limitation makes Cross Matrix Module unsuitable for direct integration into DiffMatch, leading us to use GEDGNN(AGNN) as the evaluation model for AGNN instead. 

The overall performance of GEDGNN(AGNN) is presented in Table \ref{tab:ablation}. The performance of GEDGNN increased significantly by incorporating AGNN, demonstrating that AGNN effectively enhances the model’s ability to predict node matching probabilities by directly computing expressive node pair embeddings.

\begin{figure}[htbp]
    \centering
    
    \begin{subfigure}{0.49\linewidth}
        \centering
        \includegraphics[width=\linewidth]{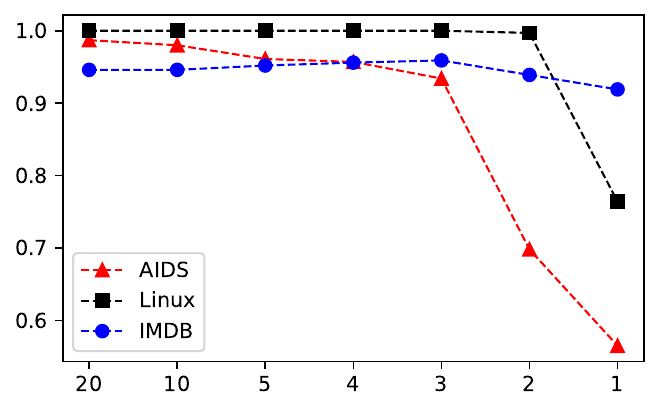}
        \caption{Accuracy vs. Denoising Steps}
    \end{subfigure}
    \hfill
    \begin{subfigure}{0.49\linewidth}
        \centering
        \includegraphics[width=\linewidth]{step_acc.pdf}
        \caption{MAE vs. Denoising Steps} 
    \end{subfigure}

    \begin{subfigure}{0.49\linewidth}
        \centering
        \includegraphics[width=\linewidth]{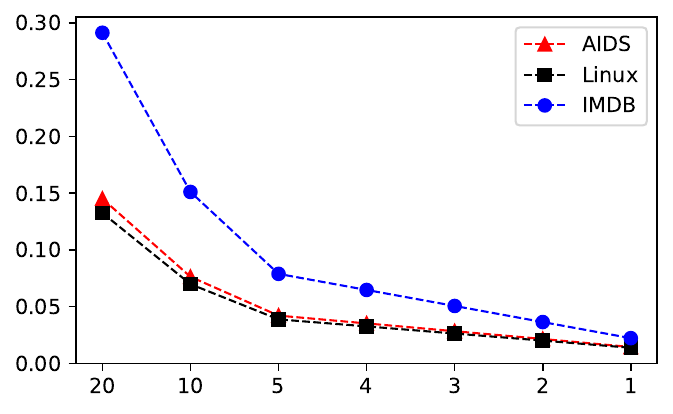}
        \caption{Time (s) vs. Denoising Steps} 
    \end{subfigure}

    \caption{Performance comparison across different reverse denoising steps during inference}
    \label{fig:diff_step}
\end{figure}

\vspace{1mm}\noindent \textbf{Varying Reverse Denoising Steps during Inference} During inference, DiffMatch denoises noisy node matching matrices through $S$ reverse steps. To assess the impact of the number of reverse denoising steps on DiffGED's performance, we evaluate DiffGED using different values of $S$, specifically $S=[20,10,5,4,3,2,1]$. FFigure \ref{fig:diff_step} presents the performance comparison across different values of $S$. The results indicate that when $S>2$,the accuracy and MAE of DiffGED do not vary a lot. However,  when $S \leq 2$, taccuracy drops significantly while MAE increases. In particular, at $S=1$, DiffGED becomes a one-shot model, suffering from the same limitations as DiffGED(w/o diffusion), leading to similarly poor performance. Moreover, when $S$ is doubled, the running time of DiffGED almost doubles as well, as the majority of its computational cost comes from denoising the node matching matrix at each reverse step. 

\begin{table}[t]
\centering
\caption{Evaluation on Node Mapping Extraction Strategy}
\label{tab:extraction}
\begin{tabular}{|c|c|c|c|c|c|c|c|c|}
\hline
Datasets & Models & MAE & Accuracy& Time(s) \\
\hline
\hline
\multirow{2}{*}{AIDS700}&DiffGED&${0.022}$&${98\%}$&$0.00043$\\
&DiffGED(Hungarian)&$0.021$&$98.1\%$&$0.0035$\\
\hline
\hline
\multirow{2}{*}{Linux}
&DiffGED&${0.0}$&${100\%}$&$0.00036$\\
&DiffGED(Hungarian)&${0.0}$&${100\%}$&$0.00345$\\

\hline
\hline
\multirow{2}{*}{IMDB}
&DiffGED&${0.937}$&${94.6\%}$&${0.00068}$\\
&DiffGED(Hungarian)&${0.918}$&${94.7\%}$&$0.00367$\\

\hline

\end{tabular}
\end{table}

\begin{figure}[t]
     \centering
     \hspace*{-0.5cm}
     \includegraphics[scale=1]{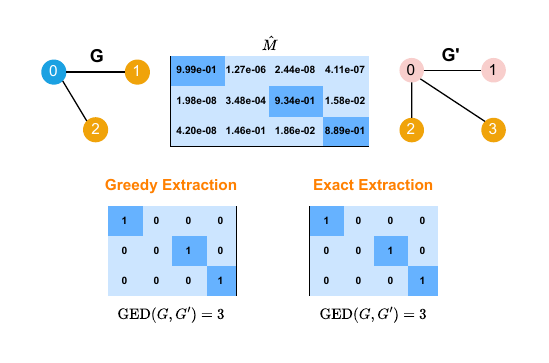}
     \caption{Greedy vs Exact Node Mapping Extraction}
     \label{fig:extraction}
\end{figure}
\vspace{1mm}\noindent \textbf{Greedy vs Exact Node Mapping Extraction} To evaluate the effectiveness and efficiency of greedy node mapping extraction, we introduce a variant model, DiffGED(Hungarian), which replaces the greedy extraction method with the exact Hungarian algorithm \cite{kuhn1955hungarian}.
As shown in Table \ref{tab:extraction}, DiffGED with greedy node mapping extraction achieves nearly identical accuracy and MAE to DiffGED(Hungarian) across all datasets, while significantly reducing the computational cost of node mapping extraction. This improvement stems from the fact that DiffMatch generates a high-quality sparse node matching matrix, where most elements in each row and column are close to 0, with only a few elements close to 1. This sparsity enables the greedy extraction method to retrieve node mappings comparable to those obtained by the exact Hungarian algorithm while being much faster. Figure \ref{fig:extraction} illustrates a small example graph pair from the AIDS dataset, where $\hat{M}$ represents the node matching matrix predicted by DiffMatch. We can see that the predicted $\hat{M}$ is both high-quality and sparse, leading to identical extracted node mappings under both the greedy and Hungarian strategies, resulting in $GED(G,G')=3$.

\section{Conclusion}
This paper presents a novel GED solver named DiffGED that recovers the optimal edit path by leveraging a generative diffusion model to generate top-$k$ node mappings. Our approach works by predicting $k$ diverse node-matching matrices simultaneously through our diffusion-based graph matching model, DiffMatch, and then extracting the top-$k$ node mappings in parallel using a greedy algorithm. Extensive experiments on real-world datasets demonstrate that our method outperforms all other hybrid approaches by generating diverse, high-quality edit paths with accuracy close to $1$, all within a short running time.

\bibliographystyle{ACM-Reference-Format}
\bibliography{main}

\appendix

\end{document}